\documentclass{article}

\usepackage{spconf,amsmath,graphicx}
\usepackage[utf8]{inputenc}
\usepackage[colorinlistoftodos]{todonotes}
\usepackage{tabularx}
\usepackage{multirow}
\usepackage{gensymb}
\usepackage{enumitem}
\usepackage{interval}
\usepackage[lofdepth,lotdepth]{subfig}
\newcolumntype{L}[1]{>{\raggedright\let\newline\\\arraybackslash\hspace{0pt}}m{#1}}
\newcolumntype{C}[1]{>{\centering\let\newline\\\arraybackslash\hspace{0pt}}m{#1}}
\usepackage[colorlinks=true,citecolor=blue]{hyperref}
\usepackage{siunitx}
\usepackage{placeins}
\usepackage{graphicx}

\makeatletter
\newcommand\footnoteref[1]{\protected@xdef\@thefnmark{\ref{#1}}\@footnotemark}
\makeatother

\newlength{\tempdima}
\newcommand{\rowname}[1]
{\rotatebox{90}{\makebox[\tempdima][c]{#1}}}

\title{STRATEGIES FOR TRAINING STAIN INVARIANT CNNS}

 \name{\parbox{\linewidth}{\centering
      Thomas Lampert$^{\star}$ \qquad Odyss\'{e}e Merveille$^{\star}$ \qquad Jessica Schmitz$^{\diamond}$ \\ \textit{Germain Forestier}$^{\dagger}$ \qquad \textit{Friedrich Feuerhake}$^{\diamond,\ddagger}$ \qquad \textit{C\'{e}dric Wemmert}$^{\star}$}\thanks{This work was supported by the SysMIFTA project (FKZ:031L-0085A). The authors are grateful to Nvidia Corporation for donating a Quadro P6000 GPU, which was used to conduct this research.}}
\address{\!\!\!\!\!\!\!$^{\star}$ICube, Université de Strasbourg, France \quad $^{\diamond}$Institute of Pathology, Hannover Medical School, Germany \\
\!\!\!\!\!\!\!$^{\dagger}$IRIMAS, Université de Haute Alsace, Mulhouse, France \quad $^{\ddagger}$University Clinic, Freiburg, Germany}

\begin{document}
\maketitle              
\begin{abstract}
An important part of Digital Pathology is the analysis of multiple digitised whole slide images from differently stained tissue sections. It is common practice to mount consecutive sections containing corresponding microscopic structures on glass slides, and to stain them differently to highlight specific tissue components. These multiple staining modalities result in very different images but include a significant amount of consistent image information. Deep learning approaches have recently been proposed to analyse these images in order to automatically identify objects of interest for pathologists. These supervised approaches require a vast amount of annotations, which are difficult and expensive to acquire---a problem that is multiplied with multiple stainings. This article presents several training strategies that make progress towards stain invariant networks.  By training the network on one commonly used staining modality and applying it to images that include corresponding but differently stained tissue structures, the presented unsupervised strategies demonstrate significant improvements over standard training strategies.
\end{abstract}


\section{Introduction}

The field of digital pathology emerged with the introduction of whole slide imaging (WSI) scanners and lead to the development of new tools for analysing histopathology slides \cite{Gurcan09}. Meanwhile, deep learning architectures were developed that outperformed most state-of-the-art algorithms in image analysis. Deep learning has thus been quickly applied to solve digital pathology problems \cite{de2018automatic,Janowczyk16,das2018multiple}. Nonetheless supervised deep learning approaches require large amounts of annotated data. Unlike classic computer vision applications, annotating histopathology slides requires in-depth knowledge and therefore can only be done by trained experts such as pathologists. Consequently, the acquisition of large annotated histopathological datasets is a complicated task \cite{Schaadt2018RoleTaskComplexity}.

Image datasets in digital pathology applications often consist of consecutive slides stained differently, each staining providing specific information on the same region of interest (see Figure \ref{fig:glomeruli_patches}, first row).  Even though differently stained slides appear very different, there is often a significant amount of consistent information between them. For example, they may both share the same counterstain (e.g.\ haematoxylin), or they may highlight different parts of the same structure.

The analysis and integration of information from different stainings is usually performed with reference to a specific organ, structure, or pattern observed in the tissue. For example, to diagnose pathologies such as breast cancer or kidney allograft rejection it is necessary to study the inflammatory micro-environment of the organ. In these cases, the relevant information is the distribution of immune cells (e.g.\ macrophages or lymphocytes) in relation to important structures of the organ, such as glomeruli for the kidney, or lobules for the breast.

To automatically perform such an analysis, the structure of interest (glo\-me\-ru\-li, lobules, etc.) should be detected in each section irrespective of the individual staining modality. This is a typical application of deep learning, assuming that annotations for each staining are available. Thus, there is a high demand for annotations in digital pathology, currently limited by the availability of experts and the complexity of the required expert knowledge \cite{Schaadt2018RoleTaskComplexity}.

This article presents several training strategies that make progress towards stain invariant networks (i.e.\ a network that can generalise between stainings). By training the network using a commonly acquired staining in renal pathology Periodic acid-Schiff reaction (PAS, a histochemical reaction to detect certain types of suger-containing tissue components) and applying the same network to images with different stainings, the presented strategies demonstrate significant improvements over standard training strategies. These can be used in settings where there is not enough data or annotations for the desired application staining. Alternatives in the literature are concerned with transfer learning or domain adaptation strategies where some take advantage of networks pretrained on huge natural image datasets \cite{bayramoglu2016}. Most of these works consider the problem of tissue variability between images that have the same staining \cite{gadermayr2016,lafarge2017} or between images from different tissues \cite{kandemir2015}. These approaches are much more complex to implement as they require specific network architectures.

The adopted CNN architecture and proposed training strategies are presented in the next section, Section \ref{sec:results} presents inter-staining glomeruli segmentation results and analysis thereof, and Section \ref{sec:conclusions} presents the study's conclusions.

\section{Method}
\label{sec:CNNarchitecture}

\subsection{CNN architecture}

The U-Net architecture \cite{Ronneberger15} is adopted as it has been proven successful in biomedical imaging \cite{litjens2017survey} and, in particular, glomeruli detection \cite{de2018automatic}. Glomeruli segmentation is framed as a two classes problem: glomeruli (patches centred on a glomeruli), and tissue (that does not contain a glomeruli, randomly sampled). The slide background (non-tissue) is removed by thresholding each image by its mean value then removing small objects from the result and closing holes. Initial experiments revealed that the same training parameter values lead to convergence irrespective of the strategy used (batch size of $8$, learning rate of $0.0001$, $60$ epochs, and the network that achieves the lowest validation loss is kept). The input patch size is $508 \times 508$ pixels.

To mitigate the lack of training samples, the use of elastic deformation \cite{Simard03} is proposed by the original authors. In addition to this (using the parameters $\sigma = 10$, $\alpha = 100$), the following augmentations are applied with an independent probability of $0.5$ (batches are augmented `on the fly'):
\begin{description}[noitemsep,topsep=0pt,leftmargin=0pt]
\item[affine:] random rotation sampled from the interval $\interval{\ang{0}}{\ang{180}}$, random shift sampled from $\interval{-205}{205}$ pixels, random magnification sampled from $\interval{0.8}{1.2}$, and horizontal/vertical flip;
\item[noise:] additive Gaussian noise\footnote{\label{note1}Using the appropriate \href{https://scikit-image.org}{scikit-image} functions.} with $\sigma \in \interval{0}{2.55}$;
\item[blur:] Gaussian filter\footnoteref{note1} with $\sigma \in \interval{0}{1}$;
\item[brightness] enhance\footnote{\label{note2}Using the appropriate \href{https://python-pillow.org}{PIL} functions.} with a factor sampled from $\interval{0.9}{1.1}$;
\item[colour] enhance\footnoteref{note2} with a factor sampled from  $\interval{0.9}{1.1}$;
\item[contrast] enhance\footnoteref{note2} with a factor sampled from $\interval{0.9}{1.1}$;
\item[stain variation] by colour deconvolution \cite{Tellez18}, $\alpha$ sampled from $\interval{-0.25}{0.25}$ and $\beta$ from $\interval{-0.05}{0.05}$.
\end{description}
These values were chosen as they result in realistic images. All samples are standardised to $\interval{0}{1}$ and normalised by the mean and standard deviation of the training set.

\subsection{Training strategies}
\label{Subsec: different approaches}

The standard approach for segmenting histopathological slides is to present the three colour channels to the network to exploit both colour and texture/structure information (referred to herein as RGB). Henceforth `structure' refers to tissue appearance (i.e.\ is not used in a biological sense).

It is hypothesised that since histopathological slides of the same staining exhibit very low colour variance when compared to natural images, a deep learning approach will rapidly learn to rely upon structures being present in specific channels (or combinations thereof), which may limit the network's ability to generalise to unseen stainings. The effects of this are already known when dealing with inter-laboratory variance of the same staining \cite{Tellez18, Gallego18}.

\begin{figure*}[!ht]
	\begin{center}
	\settoheight{\tempdima}{\includegraphics[width=2.5cm]{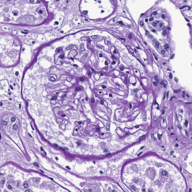}}%
	\begin{tabular}{@{}c@{ }c@{ }c@{ }c@{ }c@{ }c@{}}
	        { } & PAS & Jones H\&E & CD68 & Sirius Red & CD34
	        \\
	        \rowname{Original} &
	        \includegraphics[width=2.5cm]{img/IFTA_Nx_0011_02_glomeruli_patch_167.png} &
			\includegraphics[width=2.5cm]{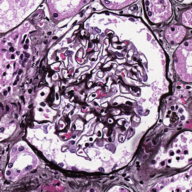} &
			\includegraphics[width=2.5cm]{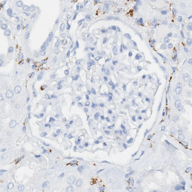} &
			\includegraphics[width=2.5cm]{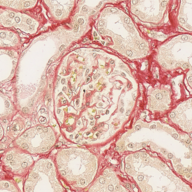} &
			\includegraphics[width=2.5cm]{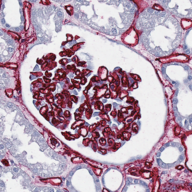}
			\\[-0.3ex]
			\rowname{Greyscale} &
			\includegraphics[width=2.5cm]{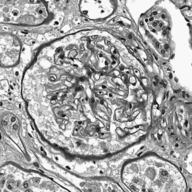}&
			\includegraphics[width=2.5cm]{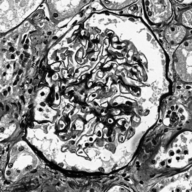}&
			\includegraphics[width=2.5cm]{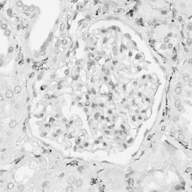}&
			\includegraphics[width=2.5cm]{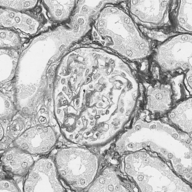}&
			\includegraphics[width=2.5cm]{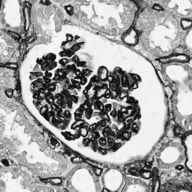}
			\\[-0.3ex]
			\rowname{Haematoxylin} &
			\includegraphics[width=2.5cm]{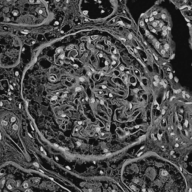}&
			\includegraphics[width=2.5cm]{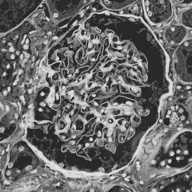}&
			\includegraphics[width=2.5cm]{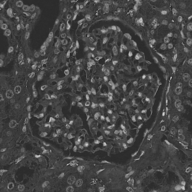}&
			\includegraphics[width=2.5cm]{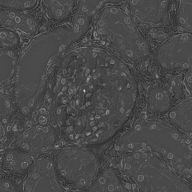}&
			\includegraphics[width=2.5cm]{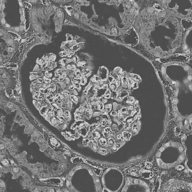}
			\\[-0.3ex]
			\rowname{Colour Transfer}&
			\includegraphics[width=2.5cm]{img/IFTA_Nx_0011_02_glomeruli_patch_167.png}&
			\includegraphics[width=2.5cm]{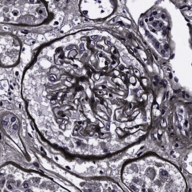}&
			\includegraphics[width=2.5cm]{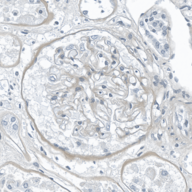}&
			\includegraphics[width=2.5cm]{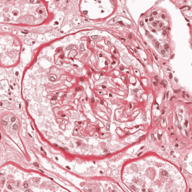}&
			\includegraphics[width=2.5cm]{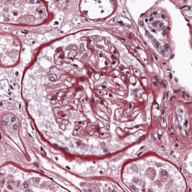}
		\end{tabular}
		\caption{Glomeruli patch examples with different transformations. Each column corresponds to a different staining, and each row a transformation. The colour transfer (4$^{\text{th}}$ row) results are applied to a PAS patch, using each staining as a target.}
		\label{fig:glomeruli_patches}
	\end{center}
\end{figure*}

Objects to be segmented, e.g.\ glomeruli, are generally easily identified between stainings as globally they exhibit the same structure and texture, see Figure \ref{fig:glomeruli_patches}. It should therefore be possible to bias the network to learn stain invariant features. This work investigates this possibility by modifying the  data presented to the network in an unsupervised manner. Borrowing domain adaptation terminology, herein the staining used for training and validation is referred to as as the \textit{source staining} (irrespective of any transformations), and the stainings to which the network are applied as \textit{target stainings}.

The simplest approach is to remove the colour component from the data (referred to herein as \textit{Greyscale}). Each sample is converted to greyscale using the formula $I = 0.2125 R + 0.7154 G + 0.0721 B$\footnoteref{note1}. The result has all colour information compressed into one channel, forcing the network to learn structural features. 

A more biologically motivated approach is to extract the haematoxylin channel from each staining by deconvolving the image using Ruifrok and Johnston's approach \cite{Ruifrok01} (referred to herein as \textit{Haematoxylin}). Haematoxylin is commonly used as a counterstain and should therefore highlight the same structures in each staining. This strategy should therefore remove stain specific information and force the network to learn features that generalise well. 

The above-mentioned strategies either compress information contained in the colour channels (Greyscale) or focus on consistent inter-staining information (Haematoxylin) with the aim of simplifying the problem. They are therefore applied during training, validation, and testing. Another approach is to preserve the colour information, but modify it in such a way that the network does not learn to rely on colour specific structure. These are therefore only applied during training and validation. A brute-force approach to this is to randomly permute the colour channels. This strategy---\textit{Channel Swap}---is simple to implement; however, the range of colours generated will be far greater than those encountered in application.

The last strategy, named \textit{Colour Transfer}, restricts this to within meaningful bounds at the expense of being more complicated. The colour profile of each training patch is replaced with that of a randomly selected staining during training with probability $\tfrac{N-1}{N}$, where $N$ is the number of stainings in the dataset. Therefore each `staining' (including the source) is presented to the network with equal probability. The colour transfer is achieved by deconvolving the patch and applying its stain concentrations to stain vectors taken from another staining. Various approaches to this exist \cite{Macenko09, Vahadane16, wemmert2013stain} (including computer vision approaches \cite{Reinhard01}), in this work we use that proposed by Macenko et al.\ due to its simplicity (automatic determination of stain vectors) and realistic output (see Figure \ref{fig:glomeruli_patches}, bottom row). It is necessary, however, to have samples of the target stainings to determine their stain vectors (this requirement can be removed by using predetermined stain vectors and Ruifrok and Johnston's approach \cite{Ruifrok01}). During training, a random patch from a random target staining is selected and the target stain vectors are determined.

N.B.\ the intention is to present plausible colour profiles to the network. This does not necessarily correspond to biologically relevant image information but is explored for any potential for the tested staining modalities. Potential limitations are addressed, e.g.\ in Figure \ref{fig:glomeruli_patches} the bright red colour (chromogenic reaction used to detect CD34, a marker of blood vessel inner lining) could be successfully transferred but it is not localised as in the original image.

\section{Experiments and results}\label{sec:results}

\subsection{Data}

Tissue samples were collected from a cohort of $10$ patients who underwent allograft nephrectomy after complete loss of function of a transplanted kidney due to chronic graft failure. These kidneys displayed several signs of rejection. The paraffin-embedded samples were cut into \SI{3}{\micro\metre} thick sections and stained with either Jones H\&E, PAS or Sirius Red, in addition to two immunohistochemistry markers (CD34 and CD68), $N=5$, using an automated staining instrument (Ventana Benchmark Ultra). Whole slide images were acquired using an Aperio AT2 scanner at $40{\times}\!$ magnification (a resolution of \SI[per-mode=fraction]{0.253}{\micro\metre\per{pixel}}). All the glomeruli in each WSI were annotated and validated by pathology experts by outlining them using Cytomine \cite{maree2016collaborative}. The dataset was divided into $4$ training, $2$ validation, and $4$ test patients. The number of glomeruli in each staining dataset was: PAS -- $666$ (train.), $580$ (valid.), $1074$ (test); Jones H\&E -- $1005$ (test); Sirius Red -- $922$ (test); CD34 -- $1013$ (test); CD68 -- $990$ (test). The training set comprised all glomeruli from the source staining training patients ($666$) and $4662$ tissue patches (to account for the variance observed in non-glomeruli tissue).

\subsection{Experiments}

For each strategy described in Section \ref{Subsec: different approaches}, a CNN was trained on the source staining (PAS) and applied to the target stainings (PAS, Jones H\&E, Sirius Red, CD34, and CD68). Test patient whole slides were segmented and precision, recall and F$_1$ score of the glomeruli were taken. To evaluate robustness, five repetitions of each experiment were performed. Table \ref{tab: quantitative results} shows the means and standard deviations of the repetitions for each measure.

The segmentation results trained on and applied to the source staining (PAS) form the controls. State-of-the-art performance \cite{de2018automatic, Bukowy18, Govind18} is achieved or surpassed and this establishes a baseline for the best achievable inter-staining performance. The segmentation results trained on the source staining and applied to the target stainings (RGB strategy) form the negative controls and, as the network essentially fails to detect anything, demonstrate the need for a transfer strategy.

The first strategy relies on compressing structural information into one band: Greyscale. This vastly increases target performance compared to RGB (with no affect on source performance) and leads to decent segmentation on Jones H\&E and CD34, which are structurally the most similar to PAS.

Next is a strategy to extract biological information that is consistent across different stainings---the haematoxylin counterstain. Despite this biological plausibility, it does not result in good results. Indeed, the haematoxylin transformed images vary greatly between the different stainings (see Figure \ref{fig:glomeruli_patches}, second row). Several factors may explain this: 1) haematoxylin's concentration relative to the primary stain may vary from one staining to another, resulting in different shades of blue and fixation amount; 2) as a counterstain, haematoxylin may become mixed with another stain in structures that are targeted by both. These result in a color mixing that is in practice not perfectly unmixed by a color deconvolution algorithm.

Although training a network using compressed structural information vastly improves inter-staining generalisation, acceptable segmentation performance is not achieved. The third strategy attempts to preserve color information but force the network not to specify the structure to particular colours. The Channel Swap strategy provides results equal to or better than the previous strategies in all but one staining (CD34), however this is achieved at the expense of stability due to the randomised colour channels.

Finally, colour variance is restricted to a more realistic range using colour transfer, giving the best results in all cases. It should be noted that the quality of the result is highly dependent on the degree of structural similarity between the target and source staining. Visually, there is a large difference between histochemical staining methods (e.g.\ H\&E, PAS, Jones) and immunohistochemistry (antibody-mediated detection of certain structures like immune cells of blood vessel inner lining) with weak blue counterstain. Therefore, some limitations of the approach are expected, and the degree of variation may explain why detection of glomeruli in the CD68 staining did not work well with PAS as source staining. 

\section{Conclusions}
\label{sec:conclusions}

It has been shown that the efficacy of simple transfer strategies depend on the degree of structural similarity between the source and target staining. The presented work dramatically improves the inter-staining segmentation performance when compared to standard training approaches, and this gain is correlated with the complexity of the strategy used. Nevertheless, it appears that a limit has been reached and strategies other than modifying training data should be the focus of future research into developing stain invariant networks.

\begin{table*}[!ht]
 \centering
 \begin{tabular}{L{2.2cm} L{1.5cm} C{2.2cm} C{2.2cm} C{2.2cm} C{2.2cm} C{2.2cm}}
  \hline \\[-2ex]
  \multirow{2}{*}{\shortstack{Training\\ Strategy}} & \multirow{2}{*}{Score} & \multicolumn{5}{c}{Test Staining}\\
  & & PAS & Jones H\&E & CD68 & Sirius Red & CD34\\
  \hline
  \hline\\ [-2ex]
  \multirow{3}{*}{RGB} & F$_1$ & \textbf{0.901} \small{(0.006)} & 0.031 \small{(0.015)} & 0.000 \small{(0.000)} & 0.015 \small{(0.012)} & 0.011 \small{(0.009)} \\
                       & precision & 0.873 \small{(0.014)} & 0.023 \small{(0.014)} & 0.372 \small{(0.292)} & 0.009 \small{(0.007)} & 0.052 \small{(0.017)}\\ 
                       & recall & 0.932 \small{(0.021)} & 0.075 \small{(0.031)} & 0.000 \small{(0.000)} & 0.063 \small{(0.067)} & 0.006 \small{(0.006)}\\ 
  \hline\\[-2ex]
  \multirow{3}{*}{Greyscale} & F$_1$ & 0.899 \small{(0.004)} & 0.599 \small{(0.080)} & 0.030 \small{(0.033)} & 0.301 \small{(0.059)} & 0.596 \small{(0.050)} \\
                       & precision & 0.867 \small{(0.013)} & 0.444 \small{(0.088)} & 0.154 \small{(0.098)} & 0.464 \small{(0.072)} & 0.703 \small{(0.058)}\\ 
                       & recall & 0.934 \small{(0.007)} & 0.944 \small{(0.007)} & 0.017 \small{(0.020)} & 0.238 \small{(0.078)} & 0.523 \small{(0.071)}\\ 
  \hline\\[-2ex]
  \multirow{3}{*}{Haematoxylin} & F$_1$ & 0.872 \small{(0.010)}  & 0.676 \small{(0.036)} & 0.002 \small{(0.001)} & 0.001 \small{(0.000)} & 0.042 \small{(0.019)} \\
                       & precision & 0.807 \small{(0.019)} & 0.609 \small{(0.096)} & 0.029 \small{(0.024)} & 0.003 \small{(0.003)} & 0.079 \small{(0.041)}\\ 
                       & recall & 0.948 \small{(0.003)} & 0.785 \small{(0.075)} & 0.001 \small{(0.001)} & 0.000 \small{(0.000)} & 0.034 \small{(0.020)}\\ 
  \hline\\[-2ex]
  \multirow{3}{*}{\shortstack{Channel\\Swap}} & F$_1$ & 0.889 \small{(0.005)} & 0.671 \small{(0.050)} & 0.106 \small{(0.111)} & 0.508 \small{(0.133)} & 0.379 \small{(0.139)} \\
                       & precision & 0.846 \small{(0.012)} & 0.659 \small{(0.063)} & 0.167 \small{(0.124)} & 0.658 \small{(0.116)} & 0.605 \small{(0.139)}\\ 
                       & recall & 0.936 \small{(0.013)} & 0.700 \small{(0.111)} & 0.090 \small{(0.108)} & 0.457 \small{(0.156)} & 0.299 \small{(0.146)}\\ 
  \hline \\[-2ex]
  \multirow{3}{*}{\shortstack{Colour\\Transfer}} & F$_1$ & 0.882 \small{(0.015)} & \textbf{0.813} \small{(0.042)} & \textbf{0.153} \small{(0.083)} & \textbf{0.739} \small{(0.070)}  & \textbf{0.709} \small{(0.028)} \\
                       & precision & 0.834 \small{(0.039)}& 0.748 \small{(0.078)} & 0.417 \small{(0.133)} & 0.742 \small{(0.106)} & 0.726 \small{(0.062)} \\ 
                       & recall & 0.938 \small{(0.015)} & 0.899 \small{(0.017)} & 0.099 \small{(0.059)} & 0.746 \small{(0.073)} & 0.706 \small{(0.089)} \\                       
  \hline
 \end{tabular}
  \caption{Quantitative results for each strategy trained on PAS (source staining) and tested on different stainings (target stainings). The numbers in parentheses are the standard deviations of the corresponding scores.}
  \label{tab: quantitative results}
\end{table*}

\FloatBarrier

%
%
%

%


\end{document}